\documentclass{article}
\usepackage{spconf,amsmath,graphicx}

\usepackage{bbding}
\usepackage{pifont}
\usepackage{wasysym}
\usepackage{amssymb}
\usepackage{array}
\usepackage{fancybox}
\usepackage{comment}
\usepackage{setspace}
\usepackage{bm}
\usepackage{multirow,color}
\usepackage{footnote}
\usepackage{enumerate}
\usepackage{amsfonts,subfigure}
\usepackage{comment,cite}

\usepackage{subfigure}
\usepackage{listliketab}
\usepackage{booktabs}
\usepackage{fancyhdr}
\usepackage{enumitem}
\newcommand\blfootnote[1]{
    \begingroup
    \renewcommand\thefootnote{}\footnote{#1}
    \addtocounter{footnote}{-1}
    \endgroup
}

\usepackage{hhline}
\usepackage{caption}
\usepackage{setspace}
\usepackage{url}
\usepackage{tabularx}

\ninept

\title{Zero-shot Composed Image Retrieval Considering \\Query-target Relationship Leveraging Masked Image-text Pairs}
\name{Huaying Zhang$^{\dag}$, Rintaro Yanagi$^{\dag}$, Ren Togo$^{\dag\dag}$, Takahiro Ogawa$^{\dag\dag}$ \ and Miki Haseyama$^{\dag\dag}$}
\address{$^{\dag}$Graduate School of Information Science and Technology, Hokkaido University, Japan\\
$^{\dag\dag}$ Faculty of Information Science and Technology, Hokkaido University, Japan\\
E-mail: \{huaying, yanagi, togo, ogawa, mhaseyama\}@lmd.ist.hokudai.ac.jp
}
\begin{document}
\small
%
\maketitle
%

\begin{abstract}
This paper proposes a novel zero-shot composed image retrieval (CIR) method considering the query-target relationship by masked image-text pairs. The objective of CIR is to retrieve the target image using a query image and a query text. Existing methods use a textual inversion network to convert the query image into a pseudo word to compose the image and text and use a pre-trained visual-language model to realize the retrieval. However, they do not consider the query-target relationship to train the textual inversion network to acquire information for retrieval. In this paper, we propose a novel zero-shot CIR method that is trained end-to-end using masked image-text pairs. By exploiting the abundant image-text pairs that are convenient to obtain with a masking strategy for learning the query-target relationship, it is expected that accurate zero-shot CIR using a retrieval-focused textual inversion network can be realized. Experimental results show the effectiveness of the proposed method.
\end{abstract}
\begin{keywords}
Multimedia information retrieval, composed image retrieval, end-to-end training, masking strategy.
\end{keywords}
\blfootnote{This work was partially supported by the JSPS KAKENHI Grant Numbers JP21H03456 and JP23K11141.}
\section{Introduction}\label{Sec.Intro}
Significant advancements in digital technology have enabled users to effortlessly construct large image databases on smartphones and personal computers~\cite{coughlin2006development}. With the escalation in data volume, efficiently retrieving desired images from these large-scale databases has grown more complex, presenting new challenges~\cite{chen2023deep}. Conventionally, image retrieval was realized on the metadata given to the images beforehand, such as keywords, object labels, text captions, etc~\cite{lu2016tag,ma2017multi}. To reduce the dependency on metadata, content-based image retrieval (CBIR) has been proposed recently~\cite{zhou2017recent}. CBIR aims at directly extracting information from the query and the candidate images for retrieval. In CBIR, reference images (image-to-image retrieval)~\cite{cao2020unifying} or descriptive texts~ (text-to-image retrieval)\cite{Song2019} are widely adopted as queries for retrieving the desired images.
\par
Image-to-image retrieval and text-to-image retrieval methods are verified as effective for users to find their desired images. However, in some situations, using only one modality of query cannot completely reflect users' intentions~\cite{Han_2017_ICCV, 718510}. This is because visual queries and text queries are good at representing different information.
For example, a unique color without a common name is easier to explain by directly showing the reference image, while the relationship between two objects can be easily understood by a single verb or preposition. In such cases, it is desirable for users to use both image query and text query to more clearly express their intentions. On the other hand, exploiting the important information and ignoring the unnecessary information contained in the image and text respectively remains a difficulty for traditional CBIR focusing on a uni-modal query~\cite{misra2017composing}.
\par
To obtain the combination of image and text query optimized for image retrieval, a novel task named composed image retrieval (CIR) has been proposed~\cite{vo2019composing}. The objective of CIR is to retrieve the target image via a query image and a query text. For example, a user may want to find an image that contains a user's pet in a specific situation. In this case, the user can use an arbitrary image of the user's pet and a text that describes the target situation to retrieve the desired image. To find a better fusion of the image-text features, recent CIR methods adopt a supervised training strategy~\cite{chen2020, Chen_2020_CVPR, Lee_2021_CVPR, Baldrati_2022_CVPR}. Specifically, they train the model on a set of $(\bm{I}^q, \bm{T}^q, \bm{I}^t)$ triplets, where $\bm{I}^q$, $\bm{T}^q$ and $\bm{I}^t$ stands for the query images, the query texts and the target images, respectively.
\par
Although supervised CIR methods using $(\bm{I}^q, \bm{T}^q, \bm{I}^t)$ triplets give a promising retrieval accuracy, collecting the training data with such triplets is a laborious process. In the collecting process, related image pairs should be collected, and then the annotators are required to observe the difference between the two images and describe the difference in text. To reduce the dependence on the $(\bm{I}^q, \bm{T}^q, \bm{I}^t)$ triplets, research on the novel task zero-shot CIR has been carried out energetically~\cite{Saito_2023_CVPR, Baldrati_2023_ICCV}. An effective approach adopted by zero-shot CIR methods is to use a textual inversion network~\cite{gal2023an} with a pre-trained visual-language model (VLM) for CIR tasks. The textual inversion network is used to map the query images into pseudo words. These pseudo words serve the same function as real words in deep learning. In the training phase, only the textual inversion network is optimized, and this process does not require $(\bm{I}^q, \bm{T}^q, \bm{I}^t)$ triplets.
On the other hand, there are still problems in training a textual inversion network. In detail, \cite{Saito_2023_CVPR} and \cite{Baldrati_2023_ICCV} merely focused on matching the images and the pseudo words, but ignored the gap between the textual inversion network training phase and the retrieval phase.
As a result, the textual inversion networks trained by these methods do not consider extracting the information necessary for retrieval, leaving space for accuracy improvements in zero-shot CIR. In addition, an end-to-end learning strategy is necessary to bridge the textual inversion network training phase and the retrieval phase.
\begin{figure*}[t]
\centering
  \centerline{\includegraphics[width=\textwidth]{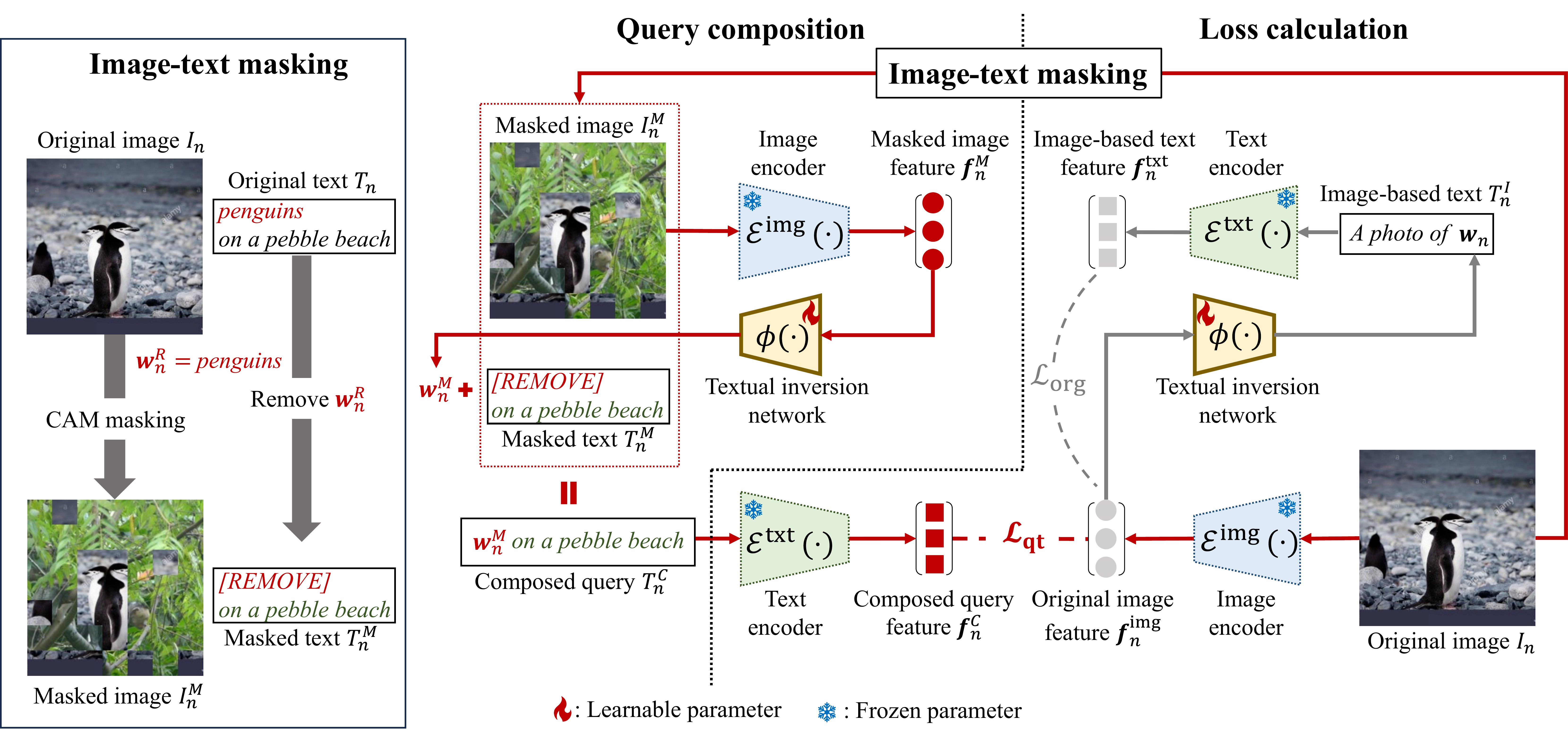}}
    \vspace{-5pt}
  \caption{Overview of our proposed method. We use the masked image-text pair as a query and the original as the target image to train the textual inversion network. The image-text masking is realized by class activation map (CAM)~\cite{Chefer_2021_ICCV}. The noun word first appears in the text is masked, and the region that is not related to the word of the image is masked by the other image in the same batch.}\medskip
  \label{Fig.Method}
  \vspace{-15pt}
\end{figure*}
\par
To address the problem that the textual inversion network is not related to the retrieval task, it is necessary to integrate the query-target retrieval relationship into training the textual inversion network. This requires the training process to be end-to-end from calculating the pseudo word to retrieving images via the VLM. Since collecting $(\bm{I}^q, \bm{T}^q, \bm{I}^t)$ triplets that are optimal for constructing the query-target relationship is laborious, we consider using only image-text pairs $(\bm{I}, \bm{T})$ that are more convenient to obtain to construct such query-target relationship.
On the other hand, visual and textual information contained in image-text pairs represents the same content, while the query image and text are complementary in CIR tasks. Thus, directly using image-text pairs does not help construct the query-target relationship. For this reason, we consider that masking the images and texts respectively can construct complementary image-text pairs. Masking is a strategy that improves the model performance by letting the model predict the masked information, which is widely adopted in the field of deep learning. Our idea is to process the image and the text so that they can contain different information from the original image-text pair and to use them to retrieve the original image. By considering the query-target relationship, it is expected that zero-shot CIR using a retrieval-focused textual inversion network can be realized.
\par
In this paper, we propose a novel zero-shot CIR method that is trained end-to-end using masked image-text pairs. We newly introduce a query-target loss by retrieving the original image by the masked image-text, so that the textual inversion network can focus on image retrieval. We first mask one word in the text and mask the parts that are not related to the word in the image, and then train the textual inversion network to minimize the distance between the composed masked image-text pair and the original image. By exploiting abundant image-text pairs that are easy to obtain, it is expected that a zero-shot CIR with higher accuracy can be realized. In addition, our method realizes an end-to-end training for zero-shot CIR.
\par
The contributions of our method are concluded as follows.
\begin{itemize}
    \item We propose a zero-shot CIR method that considers the query-target relationship in the training phase to improve the retrieval performance.
    \item We exploit the abundant image-text pairs for zero-shot CIR by applying a masking strategy on image-text pairs.
    \item We realize an end-to-end training with a lightweight textual inversion network.
\end{itemize}
\section{Zero-shot CIR with Query-target Relationship}\label{Sec.Method}
We present the proposed zero-shot CIR method in this section. Figure~\ref{Fig.Method} provides an overview of our method. The proposed method is composed of the following three steps: image-text masking, query composition, and loss calculation. We use images $I_n$ ($n = 1,\dots,N$; $N$ being the batch size) and the pair texts $T_n$ ($n = 1, \dots, N$) as one mini batch for training. First, we obtain the masked image $I^M_n$ and the masked text $T^M_n$ that represent different information respectively from the original image-text pair. Next, we compose the masked image $I^M_n$ and the masked text $T^M_n$ to obtain the composed query $T^C_n$ using a textual inversion network $\phi(\cdot)$. Intuitively, $T^C_n$ is used as a query to retrieve the original image $I_n$. Finally, we use $T^C_n$ and $I_n$ to calculate the total loss as follows:
\begin{equation}
    \mathcal{L}_{\rm total} = \alpha \mathcal{L}_{\rm qt} + \mathcal{L}_{\rm org},
\end{equation}
where $\mathcal{L}_{\rm qt}$ is the newly proposed query-target loss, $\mathcal{L}_{\rm org}$ follows the conventional zero-shot CIR method~\cite{Saito_2023_CVPR}, and $\alpha$ is the weight of $\mathcal{L}_{\rm qt}$. In our method, we use the image encoder $\mathcal{E}^{\rm img}(\cdot)$ and the text encoder $\mathcal{E}^{\rm txt}(\cdot)$ from a pre-trained cross-modal model as the backbone. In the following subsections, we explain each step of the proposed method in detail.
\subsection{Image-text Masking}\label{Subsec.2.1}
In the first phase, we conduct masking on the image-text pairs as shown in Fig.~\ref{Fig.Method}. The objective of masking is to obtain visual and textual representations from an image-text pair that are not overlapped. For simplicity, we note $T_n=\{ \bm{w}_{\rm sos}, \bm{w}_1, \dots, \bm{w}_{L_n}, \bm{w}_{\rm eos} \}$, where $L_n$ is the length of the $T_n$, and $\bm{w}_{\rm sos}$ and $\bm{w}_{\rm eos}$ are the start and the end symbol of a text. $\bm{w}_l \in [1, L_n]$ are vectorized words of $D^W$ dimension, which can be directly processed by the text encoder. First, we acquire the masked text $T^M_n$. We extract the noun words in the text $T_n$ and obtain the word $\bm{w}^R_n$ and its position that first occurs in the text automatically. We then obtain the masked text $T^M_n$ by removing the word $\bm{w}^R_n$ from $T_n$. For example, for the text \{\textit{penguins on a pebble beach}\}, the masked text will be \{\textit{[REMOVE] on a pebble beach}\}.
\par
Next, we obtain the masked image $I^M_n$. The image $I_n$ is first split into $P\times P$ patches for the following process. Then we use $\bm{w}^R_n$ to calculate the class activation map (CAM)~\cite{Chefer_2021_ICCV} of the image, noted as $M_n$. $M_n$ is a $P\times P$ matrix revealing the CAM score of each region of the image, and a higher CAM score shows that the region is more relevant to the word $\bm{w}^R_n$. Afterward, a relevant mask $M^{\rm rel}_n$ and an irrelevant mask $M^{\rm irr}_n$ are obtained by splitting $M_n$ into two parts according to a threshold $\tau$. $M^{\rm rel}_n$ and $M^{\rm irr}_n$ are 
complementary, noted as $M^{\rm irr}_n=\hat{M}^{\rm rel}_n$. Intuitively, $M^{\rm rel}_n$ masks the regions relevant to $\bm{w}^R_n$ and $M^{\rm irr}_n$ masks the regions irrelevant to $M^{\rm irr}_n$. After acquiring $M^{\rm rel}_n$ and $M^{\rm irr}_n$, we obtain the masked image $I^M_n$ as follows:
\begin{equation}
    I^M_n = I_n \odot M^{\rm irr}_n + I_m \odot M^{\rm rel}_n,
\end{equation}
where $\odot$ means element wise dot production, $n,m \in [1,\dots, N]$, and $I_m$ is an image belonging to the same batch as $I_n$. By masking $I_n$ with $I_m$ instead of simple color blocks, the completeness of the image can be preserved. Considering that complete images are used as queries in the CIR inference phase, this masking strategy is more helpful in constructing the query-target relationship in training. The masked image $I^M_n$ is expected to maintain the relevant regions to the word $\bm{w}^R_n$ in $I_n$, while the irrelevant regions are expected to be replaced by patches of another image $I_m$ in the same batch. The reason for creating such a masked image is to simulate the situation in the CIR task that query images often contain information not related to the target images.
\subsection{Query Composition}\label{Subsec.2.2}
In the second phase, we utilize the masked image $I^M_n$ and the masked text $T^M_n$ to construct the composed query $T^C_n$. The objective is to obtain a representation that can restore the full information in the image-text pair. Specifically, we first calculate the image feature $\bm{f}^M_n$ from the masked image $I^M_n$ using the image encoder $\mathcal{E}^{\rm img}(\cdot)$ as follows:
\begin{equation}
    \bm{f}^M_n = \mathcal{E}^{\rm img}(I^M_n),
\end{equation}
where the dimension of $\bm{f}^M_n$ is $D^I$. Next, we use a textual inversion network $\phi(\cdot)$ to calculate the masked pseudo word $\bm{w}^M_n$ from $\bm{f}^M_n$ as follows:
\begin{equation}
    \bm{w}^M_n = \phi(\bm{f}^M_n),
\end{equation}
where the dimension of $\bm{w}^M_n$ is equal to $D^W$. Note that the textual inversion network $\phi(\cdot)$ is a trainable lightweight $n$-layer multilayer perceptron (MLP), and the input and the output dimension of $\phi(\cdot)$ are $D^I$ and $D^W$, respectively. The pseudo word $\bm{w}^M_n$ is supposed to represent the visual information in the word format, following the conventional methods~\cite{Saito_2023_CVPR}.
\par
Next, we insert the masked pseudo word $\bm{w}^M_n$ into the position where the removed word $\bm{w}^R_n$ used to be in the masked text $T^M_n$. This enables us to obtain the composed query $T^C_n$, which is described as follows: 
\begin{align}
    T^C_n &= \{ \bm{w}_{\rm sos}, \bm{w}_1, \bm{w}_2, \dots, \bm{w}_{\rm pos}, \dots, \bm{w}_{\rm eos} \}, \\
    \bm{w}_{\rm pos} &= \bm{w}^M_n,
\end{align}
where ${\rm pos}$ is the position of which the removed word $\bm{w}^R_n$ used to be. The composed query $T^C_n$ is then utilized the query in the next step.
\subsection{Loss Calculation}\label{Subsec.2.3}
In the final phase, we calculate the total loss, which consists of two parts: the query-target loss $\mathcal{L}_{\rm qt}$ and the original loss $\mathcal{L}_{\rm org}$. We first calculate the query-target loss $\mathcal{L}_{\rm qt}$, which is newly introduced. Conventional zero-shot CIR methods usually focus on the textual inversion network to extract the visual information from the query texts and do not consider including the query-target relationship in the training phase. This is partially due to the lack of data containing queries and the corresponding targets. Our method solves this problem by utilizing the composed query obtained from Subsec.~\ref{Subsec.2.1} and Subsec.~\ref{Subsec.2.2}. Specifically, we use the composed query $T^C_n$ as the query and the original image $I_n$ as the corresponding target. First, we calculate the original image feature $\bm{f}^{\rm img}_n$ and the composed query $\bm{f}^{C}_n$ as follows:
\begin{align}
    \bm{f}^{\rm img}_n &= \mathcal{E}^{\rm img}(I_n), \\
    \bm{f}^{C}_n &= \mathcal{E}^{\rm txt}(T^C_n),
\end{align}
where $\mathcal{E}^{\rm img}(\cdot)$ and $\mathcal{E}^{\rm txt}(\cdot)$ are the image encoder and the text encoder from the pre-trained cross-modal model. The query-target loss is then calculated as follows:
\begin{align}
    L^{\rm i2t}_{\rm qt} &= -\frac{1}{N}\sum_{i\in[1,N]}{\log \frac{\exp{({\bm{f}^{\rm img}_i}^{\top}\bm{f}^{C}_i)}}{\sum_{j\in[1,N]}{\exp{({\bm{f}^{\rm img}_i}^{\top}\bm{f}^{C}_j)}}}}, \\
    L^{\rm t2i}_{\rm qt} &= -\frac{1}{N}\sum_{i\in[1,N]}{\log \frac{\exp{({\bm{f}^{C}_i}^{\top}\bm{f}^{\rm img}_i)}}{\sum_{j\in[1,N]}{\exp{({\bm{f}^{C}_i}^{\top}\bm{f}^{\rm img}_j)}}}},\\
    \mathcal{L}_{\rm qt} &= \frac{1}{2}\mathcal{L}^{\rm i2t}_{\rm qt}+ \frac{1}{2}\mathcal{L}^{\rm t2i}_{\rm qt}.
\end{align}
Here, both text-to-image and image-to-text directional losses are utilized since such a strategy is verified effective in improving the retrieval accuracy~\cite{wu2013cross}.
\begin{table*}[t]
\centering
\caption{Experimental results of PM and comparative methods on the CIRR test set. Bold indicates the best results and underline indicates the second best results. - indicates that the results are not reported in \cite{Saito_2023_CVPR} and \cite{Baldrati_2023_ICCV}.}
\label{Tab.CIRR}
\begin{tabular}{clcccc|ccc}
\hline
 & 	& \multicolumn{4}{c|}{Recall}     & \multicolumn{3}{c}{$\text{Recall}_{\text{Subset}}$}\\
\hhline{*{2}{|~}*{4}{|-}*{3}{|-}}
 Backbone &Method  &R@1 &R@5 &R@10 &R@50 &$\text{R}_{\text{Subset}}$@1 &$\text{R}_{\text{Subset}}$@2 &$\text{R}_{\text{Subset}}$@3\\
\hline\hline
\multirow[m]{2}{*}{B/32}    & PALAVRA~\cite{eccv2022_palavra_cohen} & 16.6 & 43.5 & 58.5 &84.0 &41.6 &65.3 &80.9    \\
                            & SEARLE~\cite{Baldrati_2023_ICCV} &24.0 &\underline{53.4} &\underline{66.8} &\underline{89.8} &\underline{54.9} &\textbf{76.6} &\textbf{88.2} \\
\hline
\multirow[m]{5}{*}{L/14}    & Image+Text &12.4 &36.2 &49.1 &78.2 &- &- &- \\
			            & Pic2Word~\cite{Saito_2023_CVPR} & 23.9 & 51.7 & 65.3 &87.8 &54.0 &74.7 &87.2 \\
			  	    & SEARLE-XL~\cite{Baldrati_2023_ICCV} & \underline{24.2} & 52.5 & 66.3 &88.8 &53.8 &\underline{75.0} &\textbf{88.2} \\
                            & PM & \textbf{26.1} & \textbf{55.2} & \textbf{67.5} &\textbf{90.2} & \textbf{56.0} & \textbf{76.6} & \underline{88.0}\\
                            \cline{2-9}
                            & Supervised~\cite{Baldrati_2022_CVPR} &30.3 &60.4 &73.2 &92.6 &- &- &-\\
\hline
		\end{tabular}
\end{table*}
%
\par
Next, the original loss is calculated following the conventional zero-shot CIR methods~\cite{Saito_2023_CVPR}. Specifically, the pseudo word $\bm{w}_n$ of the original image feature $\bm{f}^{\rm img}_n$ is first calculated from the textual inversion network $\phi(\cdot)$ as follows:
\begin{equation}
    \bm{w}_n = \phi(\bm{f}^{\rm img}_n).
\end{equation}
Note that $\phi(\cdot)$ is the same network as defined in Subsec.~\ref{Subsec.2.2}. Then the pseudo word $\bm{w}_n$ is used to construct the an image-based text $T^{I}_n$ \{\textit{a photo of} $w_n$\}. After that, the image-based text feature $\bm{f}^{\rm txt}_n$ is calculated by the text encoder $\mathcal{E}^{\rm txt}(\cdot)$ as follows:
\begin{equation}
    \bm{f}^{\rm txt}_n = \mathcal{E}^{\rm txt}(T^{I}_n).
\end{equation}
The original image feature $\bm{f}^{\rm img}_n$ and the image-based text feature $\bm{f}^{\rm txt}_n$ are then used to calculated the original loss $\mathcal{L}_{\rm org}$ as follows:
\begin{align}
    L^{\rm i2t}_{\rm org} &= -\frac{1}{N}\sum_{i\in[1,N]}{\log \frac{\exp{({\bm{f}^{\rm img}_i}^{\top}\bm{f}^{\rm txt}_i)}}{\sum_{j\in[1,N]}{\exp{({\bm{f}^{\rm img}_i}^{\top}\bm{f}^{\rm txt}_j)}}}}, \\
    L^{\rm t2i}_{\rm org} &= -\frac{1}{N}\sum_{i\in[1,N]}{\log \frac{\exp{({\bm{f}^{\rm txt}_i}^{\top}\bm{f}^{\rm img}_i)}}{\sum_{j\in[1,N]}{\exp{({\bm{f}^{\rm txt}_i}^{\top}\bm{f}^{\rm img}_j)}}}},\\
    \mathcal{L}_{\rm org} &= \frac{1}{2}\mathcal{L}^{\rm i2t}_{\rm org}+ \frac{1}{2}\mathcal{L}^{\rm t2i}_{\rm org}.
\end{align}
Following the assumption of the previous work~\cite{Saito_2023_CVPR, Baldrati_2023_ICCV}, the pseudo word is supposed to represent the original constantly if the text feature calculated from a generic prompt (i.e., \textit{a photo of}) plus the pseudo word has a high similarity to the corresponding image feature. In our method, we reserve this loss function to maintain the ability of $\phi(\cdot)$ to transfer the visual information of images to pseudo words.
\par
With the three steps, the zero-shot CIR method trained end-to-end that considers the query-target relationship in the training phase to improve the retrieval performance can be realized.
\begin{figure*}[!t]	
\centering
  \centerline{\includegraphics[width=0.95\textwidth]{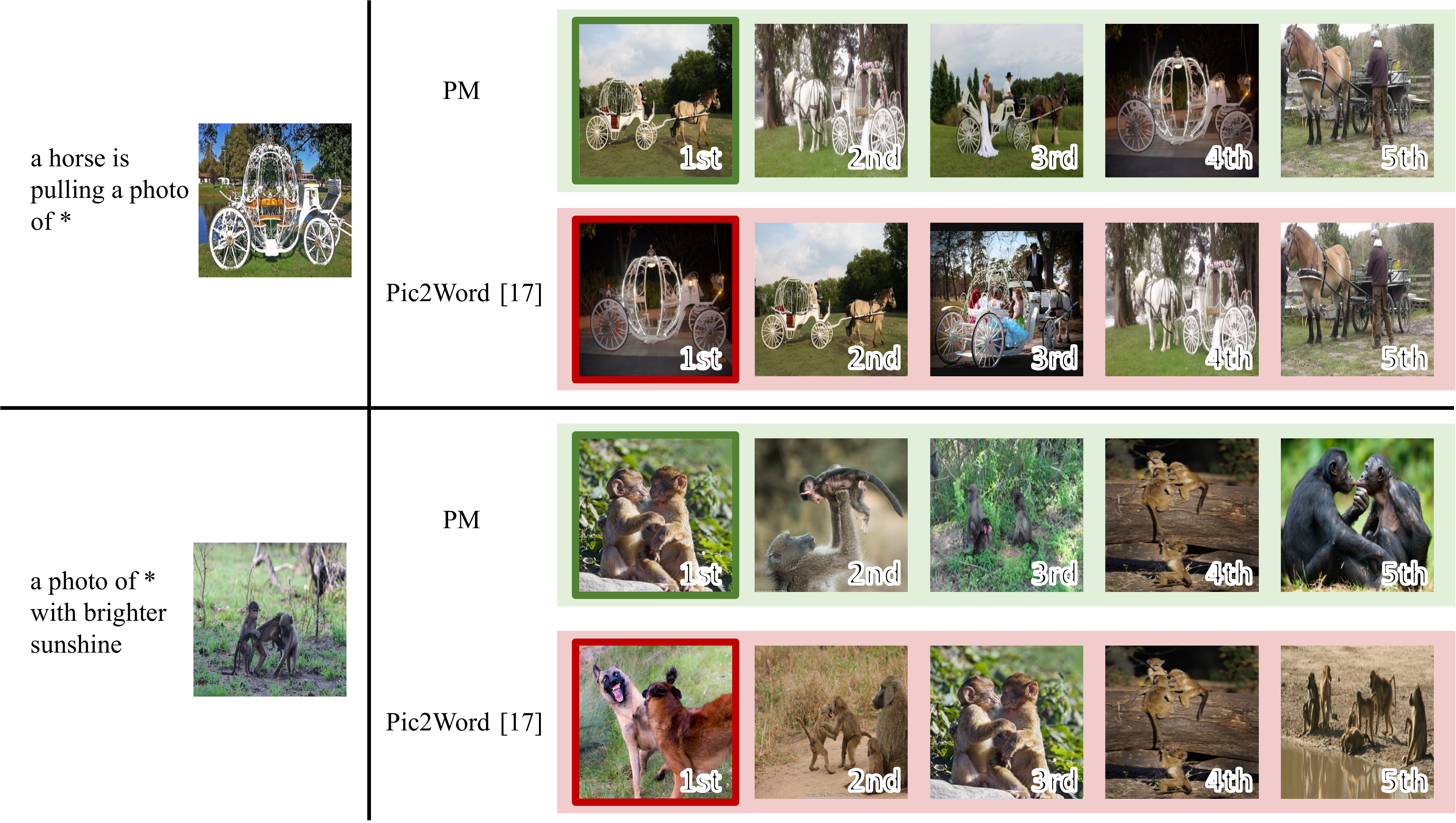}}
  \caption{Qualitative results of PM and Pic2Word~\cite{Saito_2023_CVPR}. The symbol * indicates the pseudo word generated from the query image. The image with a green frame means the ground truth image is at the top of the retrieval list, while the image with a red frame means a false image is ranked at the top.}\medskip
  \label{Fig.Qualitative}
  \vspace{-10pt}
\end{figure*}
\section{Experiments}\label{Sec:Exp}
We conducted experiments using widely adopted CIR benchmarks to evaluate the effectiveness of our method. The experimental settings and results are described in the following subsections.
\subsection{Experiment Details}\label{Subsec.3.1}
\textbf{Dataset.} For the training phase, we used 250,000 image-text pairs from CC3M published in 2018~\cite{sharma2018conceptual}. The CC3M dataset contains images and corresponding texts that describe the contents of the paired image. After the training, we evaluated the retrieval performance of our approach using a standard CIR benchmark CIRR published in 2021~\cite{Liu:ICCV2021}. We used the CIRR test set which contains $(\bm{I}^q, \bm{T}^q, \bm{I}^t)$ triplets based on real-life images for the main experiment. The number of triplets in the test set is 4,148. Note that the ground truth labels of the test set are not public, and the test is conducted on the evaluation server host by the author~\cite{Liu:ICCV2021}. In the experiment, following the previous study~\cite{Saito_2023_CVPR, Baldrati_2023_ICCV}, we utilized a query \{\textit{a photo of $\bm{w}^q_n$ , $T^q_n$}\} for each $(I^q_n, T^q_n, I^t_n) \in (\bm{I}^q, \bm{T}^q, \bm{I}^t)$ in the retrieval experiment, where $w^q_n$ is calculated by the image encoder $\mathcal{E}^{\rm img}(\cdot)$ and the textual inversion network $\phi(\cdot)$ as follows:
\begin{equation}
    \bm{w}^q_n = \phi(\mathcal{E}^{\rm img}(I^q_n)).
\end{equation}
To verify the effectiveness of our method on a domain different from the training data, we also conducted an additional retrieval experiment on a fashion domain dataset FashionIQ published in 2020~\cite{Wu_2021_CVPR}. The FashionIQ dataset contains images of clothes from three categories.
\par
\textbf{Implementation Details.} We adopted the pre-trained CLIP~\cite{radford2021learning} ViT-L/14 model as the backbone. For the image-text masking (introduced in Subsec.~\ref{Subsec.2.1}), we used the CAM~\cite{Chefer_2021_ICCV} approach based on CLIP ViT-B/32, which is a lighter model that can reduce the need for GPU memory. The threshold of CAM $\tau$ was set to 0.3. For the textual inversion network $\phi(\cdot)$, we constructed a 3-layer MLP network, with the input dimension $D^I$ and the output dimension $D^W$ equal to 768 and 768, respectively. The dimension of the middle layer was set to 3,072. The weight of the query-target loss $\alpha$ was set to 0.5. In the experiment, we adopted AdamW Optimizer~\cite{loshchilov2018decoupled}, and the initial learning rate was set to 0.0001. We trained our method end-to-end on a single RTX 6000 GPU with a batch size of 128 for 10 epochs. The training process took approximately seven hours.
\par
\textbf{Comparative Methods.} To verify the effectiveness of our proposed method (noted as PM), we compared PM with several newly proposed zero-shot CIR methods. The comparative methods we utilized are listed as follows:
\begin{enumerate}[label=(\alph*)]
\item \textbf{Image+Text}: Averaging the CLIP features of the query image and the query text as the composed query;
\item \textbf{PALAVRA}~\cite{eccv2022_palavra_cohen}: Training a textual inversion network in a two-stage method;
\item \textbf{Pic2Word}~\cite{Saito_2023_CVPR}: Training a textual inversion network only using images, the baseline of our method;
\item \textbf{SEARLE}~\cite{Baldrati_2023_ICCV}: Training a textual inversion network by a two-stage approach and considering the relationship between pseudo and real words;
\item \textbf{SEARLE-XL}~\cite{Baldrati_2023_ICCV}: Same as SEARLE but using CLIP ViT-L/14 as the backbone.
\end{enumerate}
Also, we show a result of a supervised method trained on the CIRR training set~\cite{Liu:ICCV2021}, noted as ``Supervised''. This result is considered to be the ideal retrieval performance. The experimental results of PALAVRA, SEARLE, and SEARLE-XL are cited from \cite{Baldrati_2023_ICCV}, and the results of Image+Text, Pic2Word, and Supervised are cited from \cite{Saito_2023_CVPR}, since these results are obtained under the same condition as our experiments. We classified the comparative methods as B/32 and L/14 based on the backbone they utilized.
\par
\textbf{Evaluation Metrics.} 
We utilized the commonly adopted Recall@$k$ (R@$k$) as the evaluation metric. The calculation of R@$k$ is articulated by the following equation:
\begin{equation}
      \label{equ9}
      \text{R@}k=\frac{g_k}{\gamma}.
\end{equation}
Here, $\gamma$ represents the number of queries for the test and $g_k$ stands for the number of queries where the top-$k$ retrieval images include the ground truth image. A larger R@$k$ in value is indicative of better retrieval performance. In addition, following \cite{Liu:ICCV2021}, we also used $\text{Recall}_{\text{Subset}}$ for evaluation on the CIRR test set. Specifically, CIRR also provides fully annotated 503 subsets, each of which subset contains 6 visually similar images. $\text{Recall}_{\text{Subset}}$@$k$ ($\text{R}_{\text{Subset}}$@$k$) is calculated as the ratio of samples where the ground-truth image is ranked in the list of top-$k$ image in its subset. $\text{R}_{\text{Subset}}$@$k$ is supposed to evaluate the fine-grained retrieval performance more fairly.
\subsection{Experimental Results}\label{Subsec.3.2}
\textbf{Quantitative Results.} We show the experimental results on the CIRR test set in Table~\ref{Tab.CIRR}. As shown in Table~\ref{Tab.CIRR}, our method shows an overall higher Recall than all the other zero-shot CIR methods. Remarkably, our method improves R@1 by 1.9 and R@5 by 2.7 than SEARLE-XL, which utilizes the same backbone as ours and is trained in a two-stage approach. The results of $\text{R}_{\text{Subset}}$ show a similar trend, with the $\text{R}_{\text{Subset}}$@3 of our method is slightly lower than SEARLE-XL. These results show the effectiveness of our method that utilizes masked image-text pairs to set the objective for training.
\par
In particular, we compared our method with Pic2Word~\cite{Saito_2023_CVPR}, which is the baseline of our proposed method. Although Pic2Word did not use texts, it was trained on the full 3,000,000 CC3M images with a batch size of 1,024, while ours was trained on only 250,000 image-text pairs from CC3M with a batch size of 128. Under this circumstance, our method outperforms Pic2Word by 2.2 in R@1, 3.5 in R@5, 6.9 in R@10 and 4.8 in R@50. These results also give verification to the effectiveness of using image-text pairs rather than single images since image-text pairs are not difficult to acquire. We also list out the result of the supervised method proposed in \cite{Baldrati_2022_CVPR} with an L/14 backbone. It can be seen that R@1, R@5, R@10 and R@50 of our method are only 4.2, 5.2, 5.7 and 2.4 lower than the supervised method. These results further verify the superiority of our method.
\par
\textbf{Qualitative Results.} Figure~\ref{Fig.Qualitative} presents the qualitative retrieval results of our proposed method and Pic2Word obtained on the CIRR validation set. We compared the results of the two methods because our proposed method is constructed based on Pic2Word. In the first example of Fig.~\ref{Fig.Qualitative}, we can see that our proposed method successfully retrieved the image of a horse pulling a carriage which is visually similar to the query image. On the other hand, Pic2Word gave the highest retrieval score to an image that shows a similar carriage but does not contain the information in the text query. In the second example, the image at the 1st place illustrates two monkeys under brighter sunshine than the query image. However, Pic2Word retrieved an image that contains two dogs instead of monkeys at 1st place. These qualitative results further indicate that our method that utilizes masked image-text pairs for training a textual inversion network is effective for the CIR task.
\begin{table}[t]
\centering
\caption{Results of PM on the CIRR validation set when $\tau$ varies.}
\label{Tab.Ablation}
\begin{tabular}{lccccc}
\hline
 & R@1 & R@5 & R@10 & R@50 \\
 \hline\hline
$\tau=0.2$ & 21.4 & 49.8 & 64.6 & 88.1\\
$\tau=0.3$ & \textbf{25.9} & \textbf{57.4} & \textbf{70.4} & \textbf{91.4}\\
$\tau=0.4$ & 25.7 & 56.1 & 68.9 & 89.6 \\
 \hline
 \end{tabular}
\end{table}
\begin{table*}[t]
\centering
\caption{Experimental results of PM and comparative methods on the FashionIQ test set. Bold indicates the best results and underline indicates the second best results. - indicates that the results are not reported in \cite{Saito_2023_CVPR} and \cite{Baldrati_2023_ICCV}.}
\label{Tab.Fashion}
\begin{tabular}{clcccccccc}
\hline
 & 	& \multicolumn{2}{c}{Shir}     & \multicolumn{2}{c}{Dress} & \multicolumn{2}{c}{Toptee} & \multicolumn{2}{c}{Average}\\
\hhline{*{2}{|~}*{2}{|-}|*{2}{|-}|*{2}{|-}|*{2}{|-}}
 Backbone &Method  &R@10 &R@50 &R@10 &R@50 &R@10 &R@50 &R@10 &R@50\\
\hline\hline
\multirow[m]{2}{*}{B/32}    & PALAVRA~\cite{eccv2022_palavra_cohen} & 21.5 & 37.1 & 17.3 &39.9 &20.6 &38.8 &19.8 &37.3    \\
                            & SEARLE~\cite{Baldrati_2023_ICCV} &24.4 &41.6 &18.5 &39.5 &25.7 &46.5 &22.9 &42.5 \\
\hline
\multirow[m]{5}{*}{L/14}    & Image+Text &16.3 &33.6 &21.0 &34.5 &22.2 &39.0 &19.8 &35.7 \\
			            & Pic2Word~\cite{Saito_2023_CVPR} & 20.0 & 40.2 & 26.2 &43.6 &27.9 &\underline{47.4} &24.7 &43.7 \\
			  	    & SEARLE-XL~\cite{Baldrati_2023_ICCV} & \underline{26.9} & \textbf{45.6} & \underline{20.6} &\textbf{43.1} &\textbf{29.3} &\textbf{50.0} &\underline{25.6} &\textbf{46.2} \\
                            & PM & \textbf{27.1} &\underline{43.8} &\textbf{21.4} &\underline{41.7} &\underline{28.9} &47.3 &\textbf{25.8} &\underline{44.2}\\
                            \cline{2-10}
                            & Supervised~\cite{Baldrati_2022_CVPR} &30.3 &54.5 &37.2 &55.8 &39.2 &61.3 &35.6 &57.2\\
\hline
		\end{tabular}
\end{table*}
\begin{figure*}[!t]	
\centering
  \centerline{\includegraphics[width=0.83\textwidth]{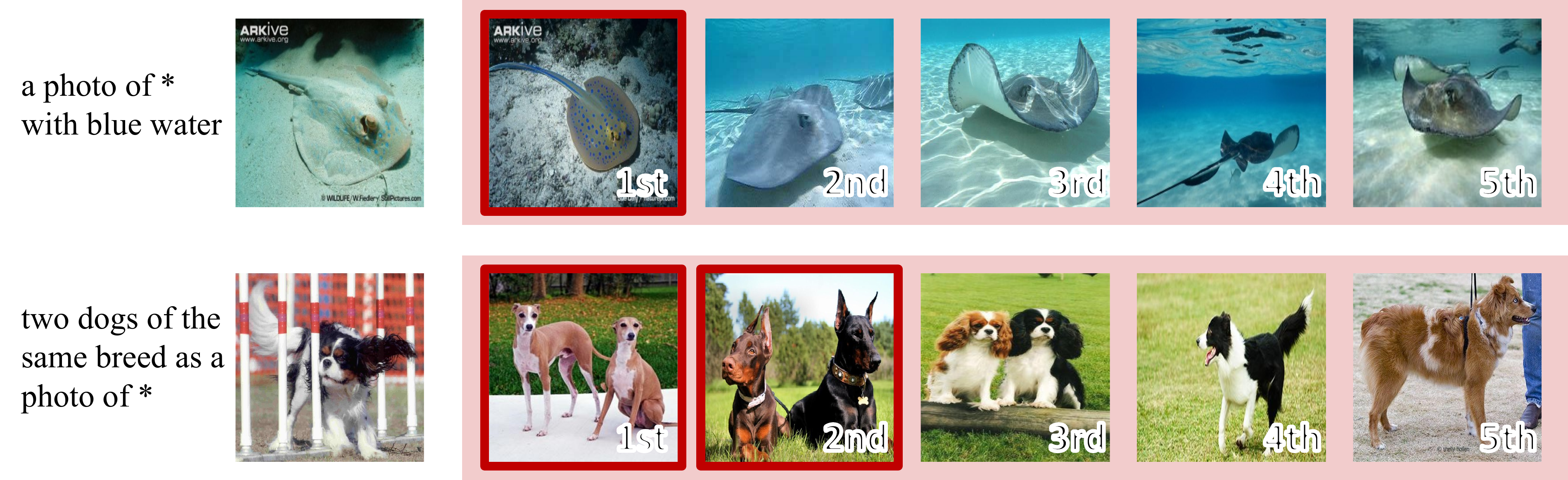}}
  \caption{Failure examples of PM. The symbol * indicates the pseudo word generated from the query image. The image with a red frame means the retrieved image is biased to the text query.}\medskip
  \label{Fig.QualitativeFailure}
  \vspace{-10pt}
\end{figure*}
%
\subsection{Additional Experiment}
\textbf{Ablation Study.} We conducted the ablation study on the hyperparameter $\tau$, which is the CAM score threshold in the range of $[0, 1]$. Since $\tau$ is a key hyperparameter of our masking strategy in the proposed method, it is vital to study its influence on retrieval performance. As is introduced in Subsec.~\ref{Subsec.2.1}, $\tau$ is used to determine the masked region of the image $I_n$ related to the removed word $\bm{w}^{R}_n$. Intuitively, the larger $\tau$ is, the more regions in the image $I_n$ are masked. In the extreme case, when $\tau$ is 0, all the regions of the image are considered relevant to $\bm{w}^{R}_n$, and thus the image would not be masked consequently. On the contrary, when $\tau$ is 1, the whole image would be masked, i.e., $I_n$ is completely substituted by $I_m$. Since the CIRR validation set is published with ground truth labels released, we conducted the ablation study on the validation set that contains 4,184 triplets. We set $\tau$ to $\{0.2, 0.3, 0.4\}$ and evaluate the proposed method under the three settings on the CIRR validation set. The experimental results are shown in Tabel~\ref{Tab.Ablation}. From Tabel~\ref{Tab.Ablation}, R@1, R@5, R@10 and R@50 all obtain a best values when $\tau=0.3$. In addition, the Recall of $\tau=0.4$ decreases compared to $\tau=0.3$. This result indicates that a mask of inappropriately large size may impose difficulties in training the textual inversion network. In addition, it is notable that the performance meets a significant drop from $\tau=0.3$ to $\tau=0.2$. We consider that it is because most images reserve their original information, which is overlapped with the text information. This can also verify the effectiveness of using masked images for training, as was discussed earlier.
\par
\textbf{Different Domain Analysis.} We also conducted an additional experiment on the FashionIQ~\cite{Wu_2021_CVPR} test set. FashionIQ contains samples in the fashion domain which is different from the domain that image-text pairs CC3M belong to. The experimental results are shown in Table~\ref{Tab.Fashion}. From Table~\ref{Tab.Fashion}, our method achieved a comparative result with SEARLE-XL. Specifically, our method obtained the best average R@10 among all the comparative methods. These results give further verification of the effectiveness of our proposed method. For the integrity of the comparison, we also posted the results achieved by the supervised method~\cite{Baldrati_2022_CVPR} for FashionIQ. The average R@10 and R@50 obtained by the proposed method are 9.8 and 13.0 lower than the supervised results, respectively.
This difference is greater than that in the CIRR dataset, which is 5.7 for R@10 and 2.4 for R@50, as shown in Table~\ref{Tab.CIRR}. We consider it due to the wider domain gap between FashionIQ and the training set compared to the gap between CIRR and the training set.
\subsection{Limitations}
Several limitations of our proposed method should be discussed. First, we illustrate several failure retrieval examples in Fig.~\ref{Fig.QualitativeFailure}. In the first example, the 1st retrieved image shows a blue-spotted ray with blue dots more obvious but does not pay attention to blue water. For the second failure case, our method focused more on \textit{dogs of the same breed} instead of \textit{the same breed as a photo of *}. These failures indicate that our method might pay more attention to the query text than the query image in some cases. We consider this due to the unbalanced masking rate of images and texts in the training phase, for only one word is masked in a text while the region masked in the image is quite large. A more reasonable masking strategy should be researched in the future. Second, our method is still trained on a dataset consisting of human-annotated image-text pairs. For future works, we will implement our proposed method on image-text pairs that are generated automatically, e.g., by using a captioning model, and verify the effectiveness of our method.
\section{Conclusion}\label{Sec.Conclusion}
In this paper, we have proposed a new zero-shot CIR method that is trained end-to-end using masked image-text pairs. By introducing the query-target loss of retrieving the original image by the masked image-text, the textual inversion network can extract useful information for retrieval. Experimental results demonstrate that our method outperforms the other zero-shot CIR methods on the CIRR dataset. For future works, we will improve the masking strategy and verify the effectiveness of our method on automatically generated image-text pairs.
\bibliographystyle{IEEEbib}
\bibliography{ref_icip2024}

\end{document}